\documentclass{svproc}
\usepackage{url}

\usepackage{amsmath,amssymb,amsfonts}
\usepackage{algorithm}
\usepackage{algorithmic}
\usepackage{graphicx}

\begin{document}
\mainmatter              % start of a contribution
\title{VELC: A New Variational AutoEncoder Based Model for Time Series Anomaly Detection}
\titlerunning{VELC}  % abbreviated title (for running head)
\author{Chunkai Zhang\and
	Shaocong Li\and
	Hongye Zhang\and
	Yingyang Chen\\}
%
%\authorrunning{Ivar Ekeland et al.} % abbreviated author list (for running head)
\authorrunning{CK. Zhang et al.}
%Craig Chambers, Kim B. Bruce, and Elisa Bertino}
%
\institute{Department of Computer Science and Technology\\Harbin Institute of Technology, Shenzhen\\Shenzhen, China,\\
\email{ckzhang@hit.edu.cn,$\{$lishaocong0327,
	zhanghongyip,
	yingyang\_chen$\}$}@163.com\\ 
}

\maketitle              % typeset the title of the contribution

\begin{abstract}
Anomaly detection is a classical but worthwhile problem, and many deep learning-based anomaly detection algorithms have been proposed, which can usually achieve better detection results than traditional methods. In view of reconstruct ability of the model and the calculation of anomaly score, this paper proposes a time series anomaly detection method based on Variational AutoEncoder model(VAE) with re-Encoder and Latent Constraint network(VELC). In order to modify reconstruct ability of the model to prevent it from reconstructing abnormal samples well, we add a constraint network in the latent space of the VAE to force it generate new latent variables that are similar with that of training samples. To be able to calculate anomaly score in two feature spaces, we train a re-encoder to transform the generated data to a new latent space. For better handling the time series, we use the LSTM as the encoder and decoder part of the VAE framework. Experimental results of several benchmarks show that our method outperforms state-of-the-art anomaly detection methods.
% We would like to encourage you to list your keywords within
% the abstract section using the \keywords{...} command.
\end{abstract}
\section{Introduction}
Anomalies often represent serious situation, unusual events or failures in many fields, such as abnormal information or cyberattack in network \cite{Onat2017An}, credit card fraud in finance\cite{Pozzolo2018Credit}, sensor anomaly in industrial field, optical coherence tomography (OCT) in medical image  \cite{Schlegl2017Unsupervised}. Anomaly detection is to find different patterns in the data which often contain important information, and these patterns are not caused by random deviations.  

According to whether the labels are used in training phase, anomaly detection algorithms can be categorized as supervised and unsupervised. The supervised method is little suitable for real data, because labeling data usually requires expert knowledge and much time cost, furthermore, the abnormal data are usually unbalanced with various kinds, and the characteristics of anomalous data may be unknown \cite{chalapathy2019deep}.
Many scholars have studied the traditional unsupervised anomaly detection methods based on the assumption that anomalies are minority class and different, such as the distance-based, density-based, angle-based methods \cite{Bayer2015Learning,pham2012near,pham2018l1}. However, the accuracy of similarity calculation, the computational complexity and accuracy of division will be greatly deteriorated due to the high-dimensionality and the large amount of data \cite{chalapathy2019deep}. Although the complexity of calculation can be reduced through simple dimensionality reduction, the detailed information will be ignored \cite{Sathe2018Subspace,Kim2018DeepNAP}. Besides, the anomalies may unknown, it means that we don’t know the percentage of anomalies in advance.

To account for this challenge, lots of unsupervised anomaly detection methods based on deep learning model are designed. One kind of method is based on the prediction error between the input samples and the prediction results or other outputs of the model to detect anomalies. \cite{malhotra2015long} used the distribution of the prediction errors of LSTMs to compute anomaly scores. \cite{sabokrou2015real} trained an autoencoder using the normal data and detects anomalies using the output of the model's hidden layer and the SSIM feature of the test samples. \cite{zhou2017anomaly} proposed RAD which combine robust principal component analysis (rPCA) with autoencoder(AE), and this model regarded the noise matrix in rPCA as an anomalous data matrix. 
The other kind of method is based on the reconstruction error of the Generative Model to detect anomalies. It commonly trains the Generative Model with the normal data in the training phase, so the model has a small reconstruction error for the normal samples but a large reconstruction error for the abnormal samples in the test phase. \cite{Bayer2015Learning} used the variational inference and RNNs to model time series data and introduced stochastic recurrent networks (STORNs), which were subsequently applied to anomaly detection in robot time series data \cite{S2016Variational}. \cite{an2015variational} proposed a method based on a VAE and introduced a novel probabilistic anomaly score that calculated the probability of the sample in the distribution corresponding to the model output parameters. \cite{zong2018deep} proposed DAGMM framework which used a deep autoencoder and a Gaussian mixture model (GMM) to model the data. The model first used the auto-encoder to reduce the dimension, and then took the compressed data and reconstruction error as a new feature. \cite{Schlegl2017Unsupervised} presented the AnoGAN framework, which trained GAN using normal data. In the test phase, AnoGAN defined a function for each test sample to find the most similar sample that the model can generate, and used the $L2$ distance between the input samples and the generated samples to detect anomalies. \cite{Zenati2018Adversarially} proposed Adversarial Learned Anomaly Detection (ALAD) based on bi-directional GANs, which added an additional discriminator in the latent space and used the adversarial learned features to detect anomalies. This model significantly improved the detection speed of GAN-based anomaly detection methods. 

Compared to the predictive model, the Generative Model can model the distribution of training data rather than the training data itself, so the strong generalization and modeling ability of the generative model usually generate some data that are similar to but different from the training data \cite{radford2015unsupervised,zhu2017unpaired}. Besides, it can compress high dimensional input samples and obtain a low-dimensional representation of the data, in such a low-dimensional representation, we can more easily distinguish the normal data and abnormal data. Although in the case of data augmentation or generation, the strong modeling and generalization capabilities of generative model will not negatively affect the output of the model, in the case of anomaly detection based on reconstruction errors of generative model, the model's strong generalization ability may make the model not only better reconstruct normal data, but even well with the abnormal data \cite{gong2019memorizing}.

In this paper, we propose a novel time series anomaly detection based on VAE with re-Encoder and Latent Constraint network, named as VELC. This model adds a re-encoder in the architecture of VAE to obtain new latent vectors, and this more complex architecture can optimize the reconstruction error both in the original space and latent space to accurately model the normal samples. Besides, it can compute the anomaly score both in the two feature spaces (original space and the latent space), which has higher accuracy than only in the original space. In addition, in order to prevent the model from reconstructing some untrained abnormal samples well, we add a constraint network in the latent space of the original VAE to force it generate new latent variables that are similar with that of training samples, which can balance the model's ability to distinguish between normal and abnormal data. During the training phase, the constraint network is trained with original VAE networks simultaneously, trying to extract the characteristics of the latent vectors of the training (normal) data. During the test phase, this network maps a latent vector obtained from the latent distribution of VAE to a new latent vector, making new latent vector similar to the latent vector corresponding to the training data. The original VAE model cannot model time series well \cite{munir2018deepant} because time series is usually high dimensional and has the complex temporal correlations, so we use the LSTM as the encoder and decoder part of the VAE framework to model the normal time series. 

The rest of this paper is organized as follows. Section II reviews the related work. Section III introduces the VELC framework and illustrates the loss function, and how to calculate the
anomaly score. Section IV shows the experimental results that our method outperforms other state-of-the-art method on several benchmarks.

\section{Related work}
Our popersed method is based on the architecture of the Variational AutoEncoder, and uses LSTM to model the normal time series, so we briefly introduce VAE and LSTM in this part. 

\subsection{VAE}
Variational AutoEncoder(VAE) \cite{kingma2013auto}, is an unsupervised deep learning Generative Model, which can model the distribution of the training data. It comes from the Bayesian inference, and consists of an encoder, latent distribution, and a decoder. The principle is a simple distribution (such as a Gaussian distribution) with known parameters and superimposable characteristics  can theoretically fit any distribution by combining with neural networks.

\iffalse
\begin{figure}[!h]
	\centering
	\includegraphics[width=3in]{pic/VAE.png}
	\caption{ The model pipeline of the VAE, E is the encoder, $\mu$ and $\sigma$ are the latent distribution parameters and D is the decoder}
	\label{fig_papr}
\end{figure}
\fi

The model's forward propagation process is as follows: the input sample $\textit{X}$ passes through the encoder to obtain parameters of the latent space distribution. The latent variable $\textit{z}$ is obtained from sampling in the current distribution, then $\textit{z}$ is used to generate a reconstructed sample through the decoder. After a series of derivation, simplification and variational inference, the loss function of VAE can be written as:

\begin{equation}
\begin{aligned}
L_{vae}(E,D) = \left \| X - X_{recon} \right \|_{2} + \frac{1}{2} \sum_{i=1}^{z_{dim}}[(\mu _{i}^{2} + \sigma_{i}^{2}) - 1 - log(\sigma _{i}^{2}))]
\end{aligned}
\end{equation}

In the forward propagation of a VAE, there is a step of sampling from the distribution. Obviously, the ``sampling'' process is not differentiable, so a method called "re-parameter" is used in VAE. Due to the superposition of the Gaussian distribution, the samples after calculation are equivalent to the sampling from the distribution corresponding to the specific distribution parameters. For the entire model, the sample obtained from the standard normal distribution can be regarded as a constant, and the calculation process for this constant is differentiable, so that the entire model can perform normal back propagation.

\begin{equation}
\varepsilon \sim N(0, 1) \quad z = \mu +\varepsilon *\sigma \quad \rightarrow \quad z \sim N(\mu , \sigma )
\end{equation}

The main difference between a VAE and an autoencoder is that the VAE is a stochastic generative model that can give calibrated probabilities, while an autoencoder is a deterministic discriminative model that does not have a probabilistic foundation. This is obvious in that VAE models the parameters of a distribution as explained above \cite{an2015variational}.

\subsection{LSTM}

Long Short Term Memory networks – usually just called “LSTMs” – are a special kind of recurrent neural network (RNN), capable of learning long-term dependencies in sequence data. LSTMs are explicitly designed to avoid the long-term dependency problem. Remembering information for long periods of time is practically their default behavior, not something they struggle to learn. The structure of LSTM is a chain form of repeating a certain neural network module(cell), which is same as RNN. The difference is that the interior of each cell of the LSTM consists of four parts: forget gate, input gate, state update and output gate. Each gate has its own weight, bias, and activation functions. 

\iffalse
\begin{figure}[!h]
	\centering
	\includegraphics[width=3in]{pic/Bi-LSTM.png}
	\caption{The framework of the Bi-LSTM, it can be seen as a combination of two simple LSTM networks in different directions.}
	\label{fig_papr}
\end{figure}
\fi

In actual sequence data, the value of a  current datapoint may not only be related to the values of some previous datapoints, but also the values of some later datapoints. So scholars proposed bi-directional LSTM network, which is seen as two LSTM networks stacked on top of each other. For a certain sequence to be trained, one of the sequences is input in the forward direction and the other is in the reverse direction, and then the two results are combined into the following formula, where $O_{t}$ is the final output, $\overrightarrow{S_{t}^{1}}$and $\overrightarrow{S_{t}^{2}}$ represent the hidden layer of two simple LSTM.

\begin{equation}
o_{t}=softmax\left(V^{*}\left[\overrightarrow{S_{t}^{\prime}} ; \overrightarrow{S_{t}^{2}}\right]\right)
\end{equation}

\begin{equation}
\overrightarrow{S_{t}^{l}}=f\left(\overrightarrow{U^{l}} * X_{t}+\overrightarrow{\vec{W}^{l}} * S_{t-1}+\overrightarrow{b^{\prime}}\right)
\end{equation}

\begin{equation}
\overrightarrow{S_{t}^{2}}=f\left(U^{2} * X_{t}+\vec{W}^{2} * S_{t-1}+\overrightarrow{b^{2}}\right)
\end{equation}

\section{Proposed Method} \label{method}
In this section, we describe a novel time series anomaly detection method based on VAE with re-Encoder and latent constraint network, named VELC. Firstly, we show the pipeline of model, and then illustrate the loss function, and how to calculate the anomaly score.

\subsection{Pipeline of Model}

The aim of our work is to detect the anomalies of a time series based on the reconstruction error of generative model, and the model is trained with normal data, it means that the model will have relatively small reconstruction errors for normal data, but large reconstruction errors for abnormal data. Our model is based on VAE with re-Encoder and latent constraint network. 

\subsubsection{Architecture of VELC}

Our model consists of four parts: an encoder and decoder of original VAE, re-Encoder and the constraint network, and it can obtain better modeling capabilities for normal data. In order to extract the features of a time series, the encoder, decoder, and re-Encoder all use bidirectional LSTM networks. The pipeline of the model is shown in Fig.1. 

\begin{figure}[h]
	%\centering
	\includegraphics[width=5in]{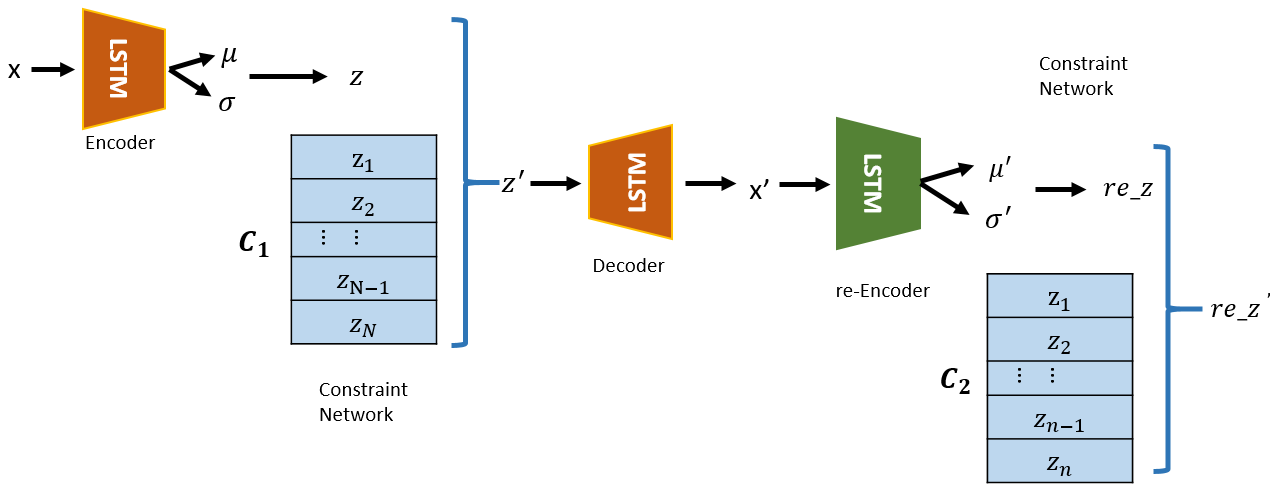}
	\caption{The network structure of VELC. Two orange blocks are the encoder and decoder layer of VAE, the green block is the re-Encoder layer and the blue part is the constraint network.}
	\label{fig_model}
\end{figure}

\subsubsection{Re-Encoder}

\ The reconstructed sample $\textit{X'}$ is passed through the re-encoder network to generate new latent space parameters $\mu$' and $\sigma$', and the new latent variables $\textit{re\_z}$ are sampled from the distribution corresponding to the new latent space parameters, and their dimension is same as the latent variables $\textit{z}$. The main purposes of the re-encoder network are as follows: The VAE with additional a re-encoder network has more parameters and complex model structure, it means that the whole model can extract more features of data, including the features of original and new latent space. Besides, similar to metric learning, it performs the accurately modeling task to normal samples directly in a certain form by optimizing the reconstruction error both in the original and latent space. Some previous model-based anomaly detection methods used the output of the network intermediate layer as feature vectors of the data and obtained better results than using the output of whole network, so the VAE with additional a re-encoder should improve the performance of model-based anomaly detection. In addition, the VAE with additional a re-encoder network can compute the anomaly score both in the two feature spaces (original space and the latent space), which has higher accuracy than only in the original space. 

\subsubsection{Constraint Network}

For anomaly detection methods based on reconstruction error of the generative model including VAE, there are two reasons that maybe result in the fail of detecting anomalies: a) the model does not reconstruct a normal sample well, which means a large reconstruction error for this normal sample; b) the model reconstructs an abnormal sample very well, which means a relatively small reconstruction error for this abnormal sample. Because the VAE is trained by the normal samples to model the distribution of training data rather than the training data itself and its objective is to reduce the reconstruction error for the normal data during the whole training phase, so it has a strong ability to reconstruct the data. But if VAE is used for anomaly detection, this strong ability may cause the model to reconstruct the untrained abnormal samples well, it means that some anomalies in the input samples are likely not detected for small reconstruction errors of them. Considering the structure of the original VAE network, we added a constraint network after sampling from the latent distribution to limit the model's ability to reconstruct abnormal data, and the constrained network can be regarded as an additional special network similar to the encoder and decoder network. Though the constrained network is a simple neural network, it is equivalent to make the decoding network more complicated. 

Inspired by the sparse autoencoder and the MemAE framework\cite{gong2019memorizing}, we have some additional rules for the constraint network. The constraint network is set to a matrix $\mathbf{C}\in (z\_dim, N)$, the number of rows of the matrix $z\_dim$ is equal to the dimension of the latent vector $\textit{z}$, the number of columns of the matrix $N$ is used as a parameter of the model, and each element of the matrix is regarded as a trained parameter by the entire model. During the training phase, the model is trained by normal samples, so each row of the model can be considered as a ``feature vector'' or ``representative vector'' of latent vectors of normal samples. 

In the phase of training or detection, for the latent variable $z$ of each sample obtained after sampling, we calculate the cosine similarity between the vector of each row of the matrix $\mathbf{C}$, which will get several representation coefficients for each $z$, denoted as vector $\vec{\mathbf{w}}$.

\begin{equation}
\mathbf{C} = (\vec{c_{1}}, \vec{c_{2}},...,\vec{c_{N}}) \quad w_{i} = \frac{z * c_{i}}{\left \| z \right \| \cdot  \left \|c_{i}  \right \|}
\end{equation}

\begin{equation}
\vec{\mathbf{w}} = (w_{1}, w_{2},..., w_{3})
\end{equation}

After obtained the vector $\vec{\mathbf{w}}$ , we normalize each element in it $\vec{\mathbf{w}} = \mathbf{w}/\left \| \mathbf{w} \right \|$. In order to prevent the latent vectors of abnormal samples from being reconstructed by a complex combination of multiple matrices $\mathbf{C}$, a sparse constraint is added to the normalized $w$. If one element of $w$ is less than a given threshold, it is set to 0, if it is greater than the threshold, it is retained. 

\begin{equation}
w_{i}^{'} = \left\{\begin{matrix}w_{i} \quad w_{i}> ths
\\ 0 \quad w_{i}\leq ths

\end{matrix}\right. \qquad \hat{\mathbf{w}} = (w_{1}^{'}, w_{2}^{'},..., w_{N}^{'})
\end{equation}

Finally, we use the linear combination of each row in the matrix $\mathbf{C}$ and $\hat{\mathbf{w}}$ as the combination coefficient to calculate a new constrained latent vector $\hat{z}$.

\begin{equation}
\hat{z} = \hat{\mathbf{w}}\mathbf{C} = w_{1}^{'}\vec{c_{1}} + w_{2}^{'}\vec{c_{2}} + ... + w_{N}^{'}\vec{c_{N}}
\end{equation}

\subsection{Loss Function}
In the training process, we train the model with normal samples, and define the loss function consisted of three parts. The first part is the loss function of original VAE as we described in the related work, which is used to reduce the $L2$ distance between the original data and the reconstructed data in the original space, and make the distribution of the model as close as possible to the distribution of the training data. We call these two losses as reconstruction loss and $KL$ loss 1.

\begin{equation}
L_{rec\_x} = \left \| X - X^{'} \right \|_{2}
\end{equation}
\begin{equation}
L_{KL\_1} = \frac{1}{2} \sum_{i=1}^{z\_dim}[(\mu _{i}^{2} + \sigma_{i}^{2}) - 1 - log(\sigma _{i}^{2}))]
\end{equation}

The second part is the loss function of the re-encoder. Following the theory of the loss of the original VAE, we designed a Loss function for the re-Encoder network, which is the same as the second term of the loss function of the original VAE.

\begin{equation}
L_{KL\_2} = \frac{1}{2} \sum_{i=1}^{z^{'}\_dim}[({\mu _{i}^{'}}^{2} + {\sigma _{i}^{'}}^{2}) - 1 - log({\sigma _{i}^{'}}^{2})]
\end{equation}

The third part is the error of the latent vector. We utilize the re-Encoder to remap the generated data to the new latent space and calculate the $L2$ distance of between the two latent vector.

\begin{equation}
L_{lat} = \left \| Z - Z^{'} \right \|_{2}
\end{equation}

For the entire model, the loss function can be described as

\begin{equation}
L_{VELC} = L_{rec\_x} + L_{KL\_1} + L_{lat} + L_{KL\_2}
\end{equation}

\subsection{Anomaly Score}
In the phase of detection, given the input samples $X=\left[x_{1}, x_{2}, \ldots, x_{N}\right] \in R^{N}$, our model can generate the new sequence data that are similar to the distribution of the training data. If the model comes across a anomaly sample, it will generate a reconstructed sample $X^{'}$ which is significantly different from the distribution of the original data, and the abnormal sample also has a large discrepancy between the latent vector $z^{'}$ obtained by encoder and the new latent vector $re\_z^{'}$ obtained by re-encoder. 

Therefore, we can design new method to calculate the anomaly score for time series. The criterion is that if the reconstruction error and the error between the two latent spaces of an input sample are large, the sample is more likely to be an abnormal sample. The anomaly score $A(x_{i})$ is defined as follows:

\begin{equation} \label{anomalyscore}
A(x_{i}) = \alpha \left \| x - x^{'} \right \|_{1} + \beta \left \| z^{'} - re\_z^{'} \right \|_{1}
\end{equation}

where $\alpha$ and $\beta$ are the parameters to constrain the penalty term, and $\alpha+ \beta=1,\alpha>0,\beta>0$. Larger anomaly score of the sample means that the generated sample deviates from the input data, in other words, the more abnormal the sample is. In order to ensure the robustness of the algorithm and to find the anomalies in different time series, we need to regularize anomaly score into the range of $[0,1]$ and set the threshold $\phi$.

\begin{equation}
A(x_i)^{\prime}=\frac{A(x_i)-m i n(A(X))}{\max (A(X))-\min (A(X))}
\end{equation}

where $A(x_i)^{\prime}$ is the regularized anomaly score. The more abnormal the data, the larger the abnormal score is and closer to 1.

\section{Experimental Results} \label{experiment}

In this section, we first describe experimental benchmark datasets and the baseline methods. Then, we conduct many experiments to show the effectiveness of our method.

\subsection{Benchmark Datasets and Baseline Methods}

% \subsubsection{Benchmark Datasets}

To illustrate the effectiveness of our proposed method, we conduct experiments on four types of time series data(Sensor, Motion, Image and Network intrusion), which are got from UCR public dataset \cite{UCRArchive2018} and UCI public data set\cite{Dua:2019}. More details can be seen in Table \ref{tab_dataset}. The size of these sequence data sets is relatively large or small, and the dimensions of the data are relatively high or low. KDD99 is the basic benchmark of network intrusion and we choose the 10\% subset as the experimental data. Due to the high proportion of outliers, ”normal” data is considered abnormal. Motion data GunPointAgeSpan involves a series action of putting the gun and aiming at a target, which translates into 150 frames per action. We extract the centroid of the hand from each frame and use its x-axis coordinate to form a time series. Image data Herring is the Otholith outlines from two classes: North sea or Thames. We convert the outlines of the image data into sequence data. We choose the minority class as anomaly class and split 20\% of the data as test data. The other test data sets are binary classification data sets or anomaly detection data sets. We use the minority class of these data sets as abnormal data, and other classes as normal data. The AR represents for the anomaly ratio in the table. 

\begin{table}[!h]
	\setlength{\tabcolsep}{5pt}
	\renewcommand\arraystretch{1}
	\centering
	\caption{The details of benchmark datasets.}
	\begin{tabular}{lllll}
		\hline
		Dataset & Data type & AR & Length &Size \\
		\hline
		KDD99 & Network & 0.15 & 121   & 494021 \\
		Arrhythmia & Sensor   & 0.20 & 274   & 452 \\
		ItalyPowerDemand & Sensor & 0.49  & 24    & 1096 \\
		TwoLeadECG & Sensor   & 0.49  & 82    & 1162 \\
		GunPointAgeSpan&Motion&0.49&150&450\\
		MoteStrain&Sensor&0.46&84&1452\\
		ToeSegmentation2&Motion&0.25&343&166\\
		Herring&Image&0.46&512&128\\
		Wafer&Sensor&0.11&152&7164\\
		ECGFiveDays&Sensor&0.20&136&884\\
		\hline
	\end{tabular}%
	\label{tab_dataset}%
\end{table}%

% \subsubsection{Baseline Methods}


To validate our method, we consider AnoGAN \cite{Schlegl2017Unsupervised}, ALAD\cite{Zenati2018Adversarially}, MLP-VAE \cite{an2015variational} and Isolation Forest \cite{Liu2009Isolation} as the baseline algorithms. Isolation Forest is a state-of-the-art non-model anomaly detection method which based on the idea of using some kinds of rule to recursively divide each dimension of the data randomly. AnoGAN uses GAN architecture to generate data that compares pixel-level differences between raw and generated data. And ALAD is an improved GAN-based anomaly detection model, mainly based on Bi-GAN \cite{donahue2016adversarial} and the ALICE \cite{li2017alice} framework. As for MLPVAE, it uses the VAE framework. For each test sample, the model detect the anomaly by sampling the model several times and calculating the probability of the test sample in a certain distribution. 

\subsection{Performance Evaluation}

Rather than precision or recall, AUC (Area Under the ROC Curve) \cite{Ling2003AUC} is the common metrics to measure performance in anomaly detection. So we use AUC to evaluate the performance of our method and baseline methods, and perform the experiments on ten datasets. 

We set parameter $w$ = 1 in training phase. The parameter of dataset KDD99 are: batch size is 50, learning rate is 1e-5, total iteration batch is 150000, the number of the row of the marix $\mathbf{C_{1}}$ and $\mathbf{C_{2}}$ (N) is 50. The parameters of dataset arrythmia are: batch size is 32, learning rate is 0.01, total iteration batch is 10000, the number of the row of the marix $\mathbf{C_{1}}$ and $\mathbf{C_{2}}$ (N) is 10. All the other UCI datasets’ parameter are: batch size is 32, learning rate is 0.005, total iteration batch is 5000, the number of the row of the marix $\mathbf{C_{1}}$ and $\mathbf{C_{2}}$ (N) is 50. And the sparse regular threshold is 0.025.

\begin{table}[h]
	\centering
	\caption{ AUC comparisons between the baseline methods and VELC. The best results are typeset in bold. }
	\setlength{\tabcolsep}{8pt}
	\begin{tabular}{lccccc}
		\hline
		Name  & OUR* &ANOGAN & ALAD & MLP-VAE & IForest \\
		\hline
		KDD99 & \textbf{0.958} & 0.887 & 0.950 & 0.622 & 0.929 \\
		Arrhythmia & \textbf{0.789} & 0.576 & 0.648 & 0.747 & 0.530 \\
		ItalyPowerDemand & \textbf{0.807} & 0.516 & 0.538 & 0.768 & 0.763 \\
		TwoLeadECG & \textbf{0.948} & 0.554 & 0.515 & 0.731 & 0.760 \\
		GunPointAgeSpan & \textbf{0.844} & 0.515 & 0.547 & 0.821 & 0.612 \\
		MoteStrain & \textbf{0.801} & 0.746 & 0.504 & 0.750 & 0.762 \\
		ToeSegmentation2 & \textbf{0.835} & 0.547 & 0.544 & 0.816 & 0.787 \\
		Herring & \textbf{0.722} & 0.488 & 0.569 & 0.627 & 0.698 \\
		Wafer &\textbf{0.967} & 0.558 & 0.587 & 0.790 & 0.847\\
		ECGFiveDays&\textbf{0.988}&0.970&0.694&0.910&0.678\\
		\hline
	\end{tabular}%
	%}
	%}
	\label{tab_auc}%
\end{table}%

From the Table\ref{tab_auc} we can see that our method has been improved on all ten datasets, increased by about 0.8\% on KDD datasets. For other data sets, the method in this paper has also improved by about 1\% to 5\% compared to other methods. It is apparent that our methods outperform the prior works for different types of sequence data. And our model is not weaker than traditional non-model machine learning methods. 

\begin{figure}[!h]
	\centering
	\includegraphics[width=3in]{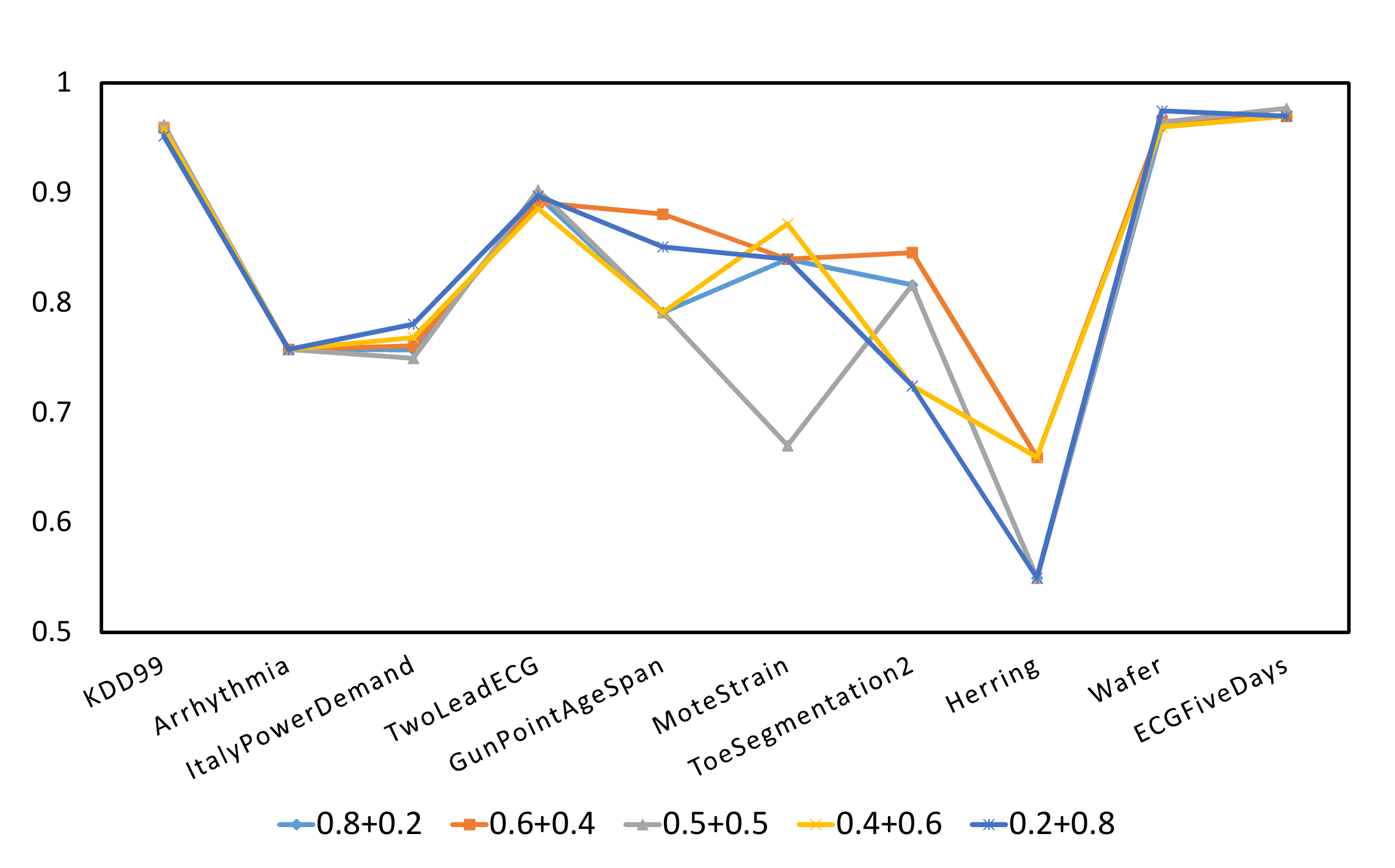}
	\caption{ Overall performance of the model based on varying parameters}
	\label{para_var}
\end{figure}

Because the anomaly score we design in Equation (\ref{anomalyscore}) is constrained by parameters $\alpha$ and $\beta$, in order to get better results, we choose different value of parameters $\alpha$ and $\beta$ varying from ($\alpha$=0.2; $\beta$=0.8) to ( $\alpha$=0.8; $\beta$=0.2). The results can be seen in Fig. \ref{para_var} that when the parameters is ($\alpha$=0.6; $\beta$=0.4), the result is better than others in average.

%\subsection{Latent Space Analysis}

\subsection{Visualization of Latent representation and Comparison of Generated Data}

%Fig.\ref{fig:lat} illustrate the lower-dimensional latent representation of the two datasets which shown with 3d scatter plot. From those plots we can see that our model can distinguish the normal and anomaly in latent space, which confirms that the model can have a better model ability of normal data. 

In order to verify the different reconstruction capabilities of the model for normal data and abnormal data, after the model training is completed, we randomly select a normal sample and an abnormal sample from the test dataset of KDD99 and ECGFiveDays, and then use the model to generate reconstructed samples, and directly compare the differences between the original and  reconstructed samples. It is shown in Fig.\ref{fig_kdd} that the reconstructed normal data is similar to the original data, and the reconstructed normal sequence is generally smoother, while the abnormal sample fluctuations are relatively large, and the generated reconstructed samples are significantly different from the original abnormal samples. It indicates that the model has ability to successfully model the distribution of normal data, and product relatively small reconstruction errors for normal data but relatively large reconstruction errors for abnormal data.

\begin{figure*}[h]
	\centering
	\includegraphics[width=5in]{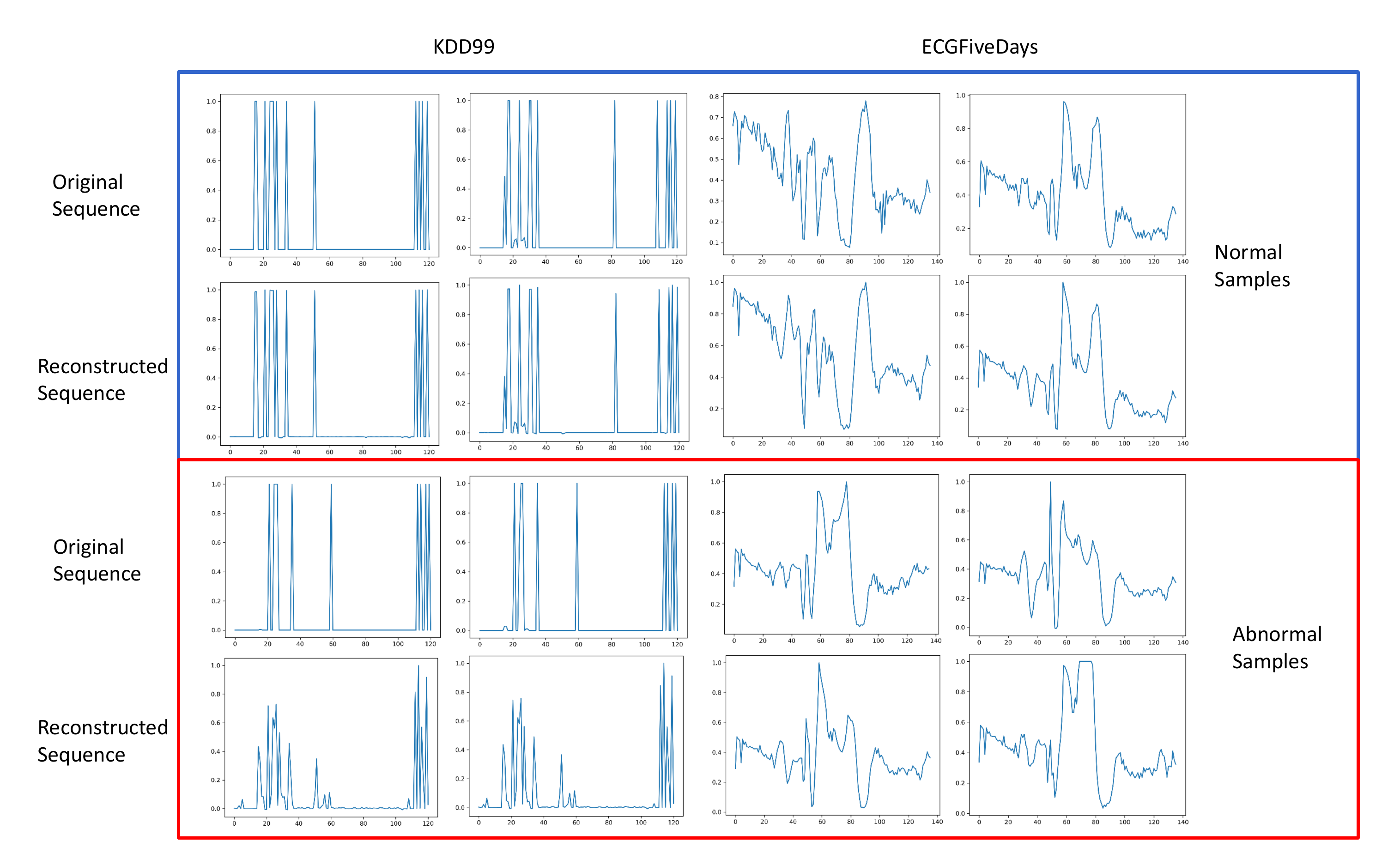}
	\caption{Comparison of reconstructed and original samples of KDD99 and ECGFiveDays respectively.}
	\label{fig_kdd}
\end{figure*}

\iffalse
\begin{figure}[h]
	\centering
	\includegraphics[width=2.6in]{pic/rec_sample3.png}
	\caption{Comparison of reconstructed and original samples of KDD99 and ECGFiveDays.}
	\label{fig_kdd}
\end{figure}
\fi

%\begin{figure*}[!t]
%		\centering
%		\includegraphics[width=4.5in]{pic/ecg5day.pdf}
%		\caption{Comparison of reconstructed and original samples of ECGFiveDays.}
%		\label{fig_ecg}
%\end{figure*}

\iffalse
\begin{figure}[h] \centering 
	\subfigure[Latent representation of Wafer.] { \label{fig:a} 
		\includegraphics[width=0.25\textwidth]{pic/lat_wafer3.png} 
	}
	\subfigure[Latent representation of Arrhythmia.] { \label{fig:b} 
		\includegraphics[width=0.25\textwidth]{pic/lat_arr.png} 
	}
	%	\subfigure[Latent representation of KDD99.] { \label{fig:c} 
	%		\includegraphics[width=0.3\textwidth]{pic/lat_kdd5.png} 
	%	}
	%	\subfigure[Latent representation of Motestrain.] { \label{fig:d} 
	%		\includegraphics[width=0.3\textwidth]{pic/lat_motestrain3.png} 
	%	} 
	%	\subfigure[Latent representation of ItalyPowerDemand.] { \label{fig:e} 
	%		\includegraphics[width=0.3\textwidth]{pic/lat_ItalyPowerDemand2.png} 
	%	}
	%	\subfigure[Latent representation of TwoLeadECG.] { \label{fig:f} 
	%		\includegraphics[width=0.3\textwidth]{pic/lat_twoleadecg.png} 
	%	} 	
	\caption{The 3D scatter plot of latent lower representation for two datdasets. The blue points are normal samples, the res triangles are the abnormal samples.} 
	\label{fig:lat} 
\end{figure}
\fi

\section{Conclusion} \label{conclusion}
In our proposed method, we design an unsupervised deep learning anomaly detection method named VELC. The model uses VAE with re-Encoder and constraint network to model the normal time series. Our results show that the model is able to detect anomalous sequence by using latent vector error and reconstruction error. Through extensive experiments, VELC outperforms state-of-the-art approaches on ten datasets. VELC’s excellent performance on each dataset also demonstrates that it is a robust model and can be applied to various applications.

\section*{Acknowledgment}

This work was supported by a grant from the Shenzhen Research Council (Grant No.GJHZ20180928155209705).


\bibliographystyle{class_splncs03_unsort}
\bibliography{mybibliography}

\end{document}